\documentclass{article} 
\usepackage{arxiv}
\usepackage{natbib}

\date{}
\renewcommand{\headeright}{}
\renewcommand{\undertitle}{}
\renewcommand{\shorttitle}{\textit{}}

\usepackage{amsmath,amsfonts,bm}

\def\eqref#1{equation~\ref{#1}}

\def\1{\bm{1}}

\DeclareMathAlphabet{\mathsfit}{\encodingdefault}{\sfdefault}{m}{sl}
\SetMathAlphabet{\mathsfit}{bold}{\encodingdefault}{\sfdefault}{bx}{n}

\usepackage{graphicx}
\usepackage{hyperref}
\usepackage{url}
\usepackage{comment}
\usepackage{multicol}
\usepackage{multirow}
\usepackage{booktabs}
\usepackage{wrapfig}

\usepackage{pgfplots}
\usetikzlibrary{pgfplots.groupplots}
\pgfplotsset{compat=1.3}
\usepackage{tikz}
\usetikzlibrary{patterns}
\usepackage{pgf-pie}

\definecolor{battleshipgrey}{rgb}{0.3, 0.3, 0.3}
\definecolor{brilliantrose}{rgb}{1.0, 0.33, 0.64}
\definecolor{americanrose}{rgb}{1.0, 0.01, 0.24}
\definecolor{jweigreen}{rgb}{0,0.45,0.24}
\definecolor{bluegray}{rgb}{0.1, 0.1, 0.4}
\definecolor{ao(english)}{rgb}{0.0, 0.5, 0.0}
\definecolor{blanchedalmond}{rgb}{1.0, 0.92, 0.8}
\definecolor{atomictangerine}{rgb}{1.0, 0.6, 0.4}
\definecolor{chocolate(web)}{rgb}{0.82, 0.41, 0.12}
\definecolor{bananayellow}{rgb}{1.0, 0.88, 0.21}
\definecolor{goldenbrown}{rgb}{0.6, 0.4, 0.08}
\definecolor{aliceblue}{rgb}{0.94, 0.97, 1.0}
\definecolor{beige}{rgb}{0.96, 0.96, 0.86}
\definecolor{babyblue}{rgb}{0.54, 0.81, 0.94}
\definecolor{camel}{rgb}{0.76, 0.6, 0.42}
\definecolor{cinnamon}{rgb}{0.82, 0.41, 0.12}

\usepackage{pgfplots}
\usetikzlibrary{pgfplots.groupplots}
\pgfplotsset{compat=1.3}
\usepackage{pifont}

\title{A Cheaper and Better Diffusion Language Model with Soft-Masked Noise}

\author{Jiaao Chen$^{\dagger\ddagger}$\thanks{Correspondence to Jiaao Chen $\textlangle$jiaaochen@gatech.edu$\textrangle$ and Aston Zhang $\textlangle$az@astonzhang.com$\textrangle$.}, Aston Zhang$^\dagger$, Mu Li$^\dagger$, Alex Smola$^\dagger$, Diyi Yang$^\diamond$\\
$^\dagger$Amazon Web Services, $^\ddagger$Georgia Institute of Technology, $^\diamond$Stanford University
}

\begin{document}

\maketitle

\setcounter{footnote}{0}

\begin{abstract}
Diffusion models that are based on iterative denoising have been recently proposed and leveraged in various generation tasks like image generation. Whereas, as a way inherently built for continuous data, existing diffusion models still have some limitations in modeling discrete data, e.g., languages. For example, the generally used Gaussian noise can not handle the discrete corruption well, and the objectives in continuous spaces fail to be stable for textual data in the diffusion process especially when the dimension is high. To alleviate these issues, we introduce a novel diffusion model for language modeling, Masked-Diffuse LM, with lower training cost and better performances, inspired by linguistic features in languages. Specifically, we design a linguistic-informed forward process which adds corruptions to the text through strategically soft-masking to better noise the textual data. Also, we directly predict the categorical distribution with cross-entropy loss function in every diffusion step to connect the continuous space and discrete space in a more efficient and straightforward way. Through experiments on 5 controlled generation tasks, we demonstrate that our Masked-Diffuse LM can achieve better generation quality than the state-of-the-art diffusion models with better efficiency.
Code is available at \url{https://github.com/amazon-science/masked-diffusion-lm}

\end{abstract}

\section{Introduction}
We present a novel diffusion method for modeling languages, Masked-Diffuse LM (language model), which uses strategic soft-masking informed by linguistic features to corrupt both the discrete and continuous space, and then iteratively denoise them back by predicting the categorical distribution. Specifically, a strategic soft-masking process is designed that gradually adds perturbation to the input text in an order from harder or more informative words to simpler or less informative words through soft-masking. As a result, the models are encouraged to recover and generate the text following an \textit{easy-first-generation} nature \cite{https://doi.org/10.48550/arxiv.2211.15089} to improve the generation structure and quality with more flexibility. Also, during the diffusion process, we directly predict the discrete token with cross-entropy loss that maps the continuous space to discrete textual space to stabilize the intermediate diffusion steps. Through our proposed Masked-Diffuse LM, the application-specific performance metrics as well as training efficiency are significantly improved over current diffusion language models based on experiments.

Our work is inspired by recent advances in diffusion models \cite{pmlr-v37-sohl-dickstein15,https://doi.org/10.48550/arxiv.2006.11239,song2021denoising,https://doi.org/10.48550/arxiv.2209.00796,https://doi.org/10.48550/arxiv.2204.06125,9878449}  that are introduced as a new generative modeling approach based on iterative denoising and have achieved high-quality generations for visual and audio modalities \cite{https://doi.org/10.48550/arxiv.2204.06125,9878449,https://doi.org/10.48550/arxiv.2205.11487,https://doi.org/10.48550/arxiv.2102.09672,https://doi.org/10.48550/arxiv.2009.09761}. 

Although these approaches have received growing attention and achieved impressive success, applying diffusion models to textual domain is still challenging and under-explored due to the discrete nature of the text (e.g., one-hot vectors) compared to continuous data like images (e.g., RGB values) \cite{https://doi.org/10.48550/arxiv.2205.14217}. A few prior works \cite{https://doi.org/10.48550/arxiv.2205.14217,https://doi.org/10.48550/arxiv.2210.08933,https://doi.org/10.48550/arxiv.2211.15029,https://doi.org/10.48550/arxiv.2107.03006,https://doi.org/10.48550/arxiv.2102.05379} that explore using diffusion models on textual data can be divided into two lines. The first is to extend diffusion models to discrete state spaces \cite{https://doi.org/10.48550/arxiv.2107.03006,https://doi.org/10.48550/arxiv.2102.05379,https://doi.org/10.48550/arxiv.2110.02037}. The second is to perform the diffusion process and its reverse process in the continuous domain and bridge the continuous and the discrete domain through embedding and rounding \cite{https://doi.org/10.48550/arxiv.2205.14217,https://doi.org/10.48550/arxiv.2211.15029}, for example, Diffusion-LM \cite{https://doi.org/10.48550/arxiv.2205.14217}. Despite the improvements, most previous works fail to leverage the linguistic features (e.g., words in sentences are with different importance) to noise the input textual data and recover it back in a more suitable way. Besides, they usually neglect or fail to adapt large pre-trained language models (PLMs) \cite{devlin2018bert, liu2019roberta, yang2019xlnet, joshi2019spanbert, sun2019ernie, clark2019electra, lewis2019bart, bao2020unilmv2, he2020deberta, raffel2020exploring}, which is an unmissable treasure in the NLP community: their adopted $k$-nearest-neighbor rounding technique that maps continuous space to discrete space cannot handle high-dimensional data in a stable and efficient way \cite{https://doi.org/10.48550/arxiv.2205.14217}.  As a result, a corruption process tailored for languages and the objective that allows efficient and straightforward discrete and continuous space transformation is in great need. Our proposed Masked-Diffuse LM realizes this extension.

To demonstrate the effectiveness of our introduced Masked-Diffuse LM, we perform experiments on E2E dataset \cite{novikova-etal-2017-e2e} and 5 controllable generation tasks \cite{https://doi.org/10.48550/arxiv.2205.14217} including Semantic Content, Parts-of-speech, Syntax Tree, Syntax Spans, and Length. We observe that our Masked-Diffuse LM can (i) achieve the state-of-the-art performances compared to recent baseline models, and (ii) allow more efficient training and inference compared to the previous Diffusion-LM.

To summarize, our contributions are: 
\begin{itemize}
    \item We introduce a strategic masking noise strategy guided by linguistic features to corrupt the textual data in diffusion models for modeling languages. 
    \item We use linear layers and cross-entropy objectives to bridge the continuous and discrete spaces in the diffusion process for efficiency and stability.
    \item We conduct experiments on different controllable generation tasks to demonstrate the effectiveness of our proposed methods compared to previous diffusion language models.
\end{itemize}

\section{Related Work}
Our work is inspired by the recent research about diffusion models, and related to or based on the work about non-autoregressive text generation and controllable generation through a plug-and-play manner.

\subsection{Diffusion Models for Language}
There has been growing attention in deep generative diffusion models, which is a latent variable generative method based on iterative denoising \cite{pmlr-v37-sohl-dickstein15,https://doi.org/10.48550/arxiv.2006.11239,song2021denoising}. Through a forward and diffusion process, diffusion models have shown state-of-the-art sample quality on generating in the continuous domain such as producing images and audio \cite{https://doi.org/10.48550/arxiv.2204.06125,9878449,https://doi.org/10.48550/arxiv.2009.09761,savinov2022stepunrolled}. Despite their huge success, it is still challenging and under-explored to adapt diffusion models to discrete domains like languages. A few recent works have modified the diffusion models for textual data. For example, discrete forward processes, such as categorical transition kernels \cite{https://doi.org/10.48550/arxiv.2110.02037}, uniform transition kernels, and absorbing kernels \cite{https://doi.org/10.48550/arxiv.2102.05379}, have been introduced. However, replacing continuous diffusion with a discrete corruption process affords some flexibility \cite{https://doi.org/10.48550/arxiv.2211.15089}. Other works have also made efforts to model text in the continuous embedding space and applied Gaussian noise uniformly to every token \cite{https://doi.org/10.48550/arxiv.2205.14217,https://doi.org/10.48550/arxiv.2211.15029}, which is closer to the settings in
previous works of diffusion models. However, they neglect the inherent linguistic features in the text (e.g., \textit{different words are playing different roles in sentences}) so the generated
text often lacks coherence \cite{https://doi.org/10.48550/arxiv.2211.15029}. Besides, the $k$-nearest-neighbor rounding technique \cite{https://doi.org/10.48550/arxiv.2205.14217} holds up the decoding and convergence speed especially when 
the vocabulary is large or the hidden dimension is high, thus limiting the potential of combining large pre-trained language models \cite{devlin2018bert, liu2019roberta, yang2019xlnet, joshi2019spanbert, sun2019ernie, clark2019electra, lewis2019bart, bao2020unilmv2, he2020deberta, raffel2020exploring}. To alleviate these issues, in our work, we introduce a linguistic-informed soft-masking process to corrupt the discrete and continuous space with structures, and then use linear projections and cross-entropy objectives to directly map the latent variables to textual data for better efficiency and generating better text.

\subsection{Non-Autoregressive Text Generation}
Most language models \cite{https://doi.org/10.48550/arxiv.2204.02311,https://doi.org/10.48550/arxiv.2005.14165} and text generation models \cite{https://doi.org/10.48550/arxiv.1706.03762,https://doi.org/10.48550/arxiv.2108.04718,chen-yang-2020-multi,chen-yang-2021-structure} follow a left-to-right autoregressive manner. However, the fixed generation order prevents the models' flexibility in editing former text based on later generation results, especially for global controllable generation settings. To overcome the limitations, non-autoregressive text modeling has been proposed \cite{ghazvininejad-etal-2019-mask,https://doi.org/10.48550/arxiv.2004.10454,gu2018nonautoregressive,saharia-etal-2020-non,savinov2022stepunrolled} through masked language models \cite{ghazvininejad-etal-2019-mask}, iterative sequence alignment \cite{saharia-etal-2020-non}, insertion and deletion \cite{gu2018nonautoregressive}, or unrolling the generation path \cite{savinov2022stepunrolled}. Our Masked-Diffuse LM achieves the non-autoregressive generation through gradually recovering the intermediate latent variables in a planned sequence from the forward process.

\subsection{Plug-and-Play Controllable Generation}
Our work is also closely related to the line of research about plug-and-play controllable generation methods \cite{yang-klein-2021-fudge,Dathathri2020Plug,krause-etal-2021-gedi-generative,liu-etal-2021-dexperts}, which modify the outputs based on extra guidance such as classifiers without changing or fine-tuning the pre-trained language models. \citet{Dathathri2020Plug} used gradients to edit the autoregressive language model's hidden representations to fulfill the control guidance. \citet{yang-klein-2021-fudge} proposed to reweight the predicted token from the language models while \cite{krause-etal-2021-gedi-generative,liu-etal-2021-dexperts} further fine-tuned a smaller LM to reweight the token predictions. In this work, we apply the gradient-based plug-and-play approach to our Masked-Diffuse LM for controllable generation by making classifier-guided gradient updates to the intermediate latent variables during the diffusion process.

\begin{figure*}[t]
\vskip 0.2in
\begin{center}
\centerline{\includegraphics[width=0.9\columnwidth]{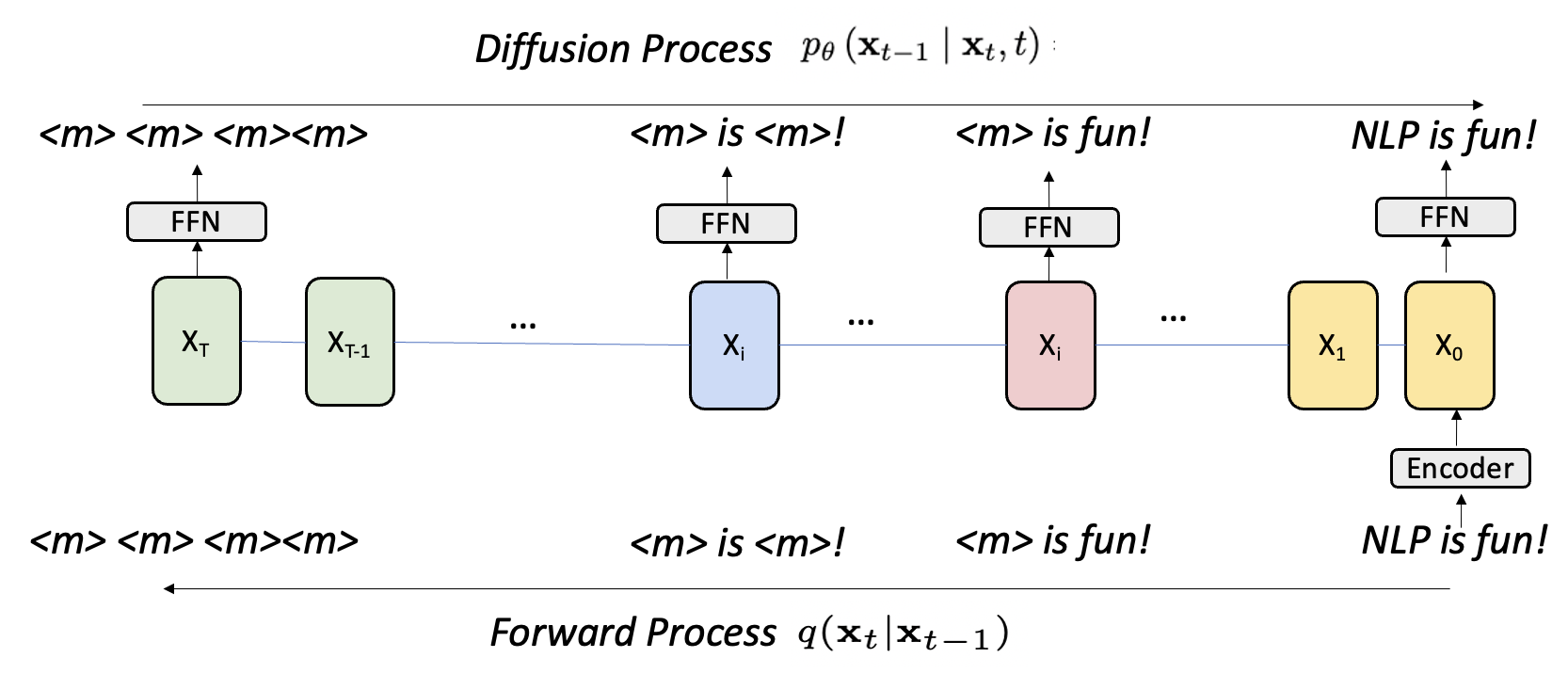}}
\caption{The overall process of our Masked-Diffuse LM. In the forward process, soft-mask is added to more informative words earlier to gradually corrupt the input text. For example, \textit{NLP} is soft-masked prior to stop words like \textit{is}. Then in the diffusion process, models learn to generate  easy words like \textit{is} first and then fill in more important words such as \textit{fun} and \textit{NLP}.} 
\label{fig:process}
\end{center}
\vskip -0.2in
\end{figure*}

\section{Background: Diffusion Models}
Diffusion models are the recent state-of-the-art deep generative models via iteratively denoising the latent variables \cite{pmlr-v37-sohl-dickstein15,https://doi.org/10.48550/arxiv.2006.11239,song2021denoising}. Basically, corruptions (usually Gaussian noise) are added to the input data distribution gradually during a forward process. Then a diffusion model is trained through learning to recover the corrupted distribution to the original input data distribution step by step. A small amount of information that is perturbed during the corresponding forward process is reconstructed in every diffusion step. The diffusion models are showing significant improvements \cite{https://doi.org/10.48550/arxiv.2204.06125,9878449,https://doi.org/10.48550/arxiv.2009.09761,savinov2022stepunrolled} as they generate the data in multiple steps, which is more stable and easier than learning to reconstruct the whole input data in a single forward pass \cite{https://doi.org/10.48550/arxiv.2211.15089} like variational autoencoders \cite{https://doi.org/10.48550/arxiv.1312.6114} and generative adversarial networks \cite{https://doi.org/10.48550/arxiv.1406.2661}.

There are usually a forward noising process and a diffusion denoising process in a diffusion model. For a given sampled input data, ${x}_0 \sim q({x}_0)$,  a Markov chain of latent variables $\{{x}_1, \cdot \cdot \cdot, {x}_T\}$ are generated in the forward noising process ($q\left({x}_t \mid {x}_{t-1}\right)$) by progressively adding a small amount of Gaussian noise to perturb the input data:
\begin{equation}
q\left({x}_t \mid {x}_{t-1}\right)=\mathcal{N}\left({x}_t ; \sqrt{1-\beta_t} {x}_{t-1}, \beta_t {I}\right),
\end{equation}
where $\left\{\beta_t \in(0,1)\right\}_{t=1}^T$ is a noise schedule controlling the amount of added noise in every step. Through the forward process,  ${x}_T$ becomes an isotropic Gaussian distribution. Note that there are no trainable parameters in the forward process.

Then a reversed diffusion process, which is learned by a parameterized model ($p({x}_{t-1}|{x}_t)$), is learned to denoise ${x}_T$ to the original data ${x}_0$:
\begin{equation}
p_\theta\left({x}_{t-1} \mid {x}_t, t\right)=\mathcal{N}\left({x}_{t-1} ; \mu_\theta\left({x}_t, t\right), \Sigma_\theta\left({x}_t, t\right)\right),
\end{equation}
where $\mu_\theta(.)$ and $\Sigma_\theta (.)$ are the learned model that can be implemented by a
U-Net \cite{https://doi.org/10.48550/arxiv.1505.04597} or a Transformer \cite{vaswani2017attention}. 

The diffusion model is trained to maximize the marginal likelihood of $\log p_\theta({x}_0)$ and we manage to minimize the variational lower bound \cite{https://doi.org/10.48550/arxiv.1503.03585} in practice:
\begin{equation}
\begin{aligned}
& \mathcal{L}_{\mathrm{vlb}}=\mathbb{E}_q\left[D_{\mathrm{KL}}\left(q\left({x}_T \mid {x}_0\right) \| p_\theta\left({x}_T\right)\right)\right] \\
& +\mathbb{E}_q\left[\sum_{t=2}^T D_{\mathrm{KL}}\left(q\left({x}_{t-1} \mid {x}_t, {x}_0\right) \| p_\theta\left({x}_{t-1} \mid {x}_t, t\right)\right)\right] \\
& -\log p_\theta\left({x}_0 \mid {x}_1\right).
\end{aligned}
\end{equation}

However, this objective is usually unstable and requires many optimization tricks to stabilize. Thus, we follow \citet{https://doi.org/10.48550/arxiv.2006.11239} to expand and reweight
each KL-divergence term in $\mathcal{L}_{\mathrm{vlb}}$ and obtain a mean-squared error ($L_2$) loss:
\begin{equation}
\mathcal{L}_{\text {diffuse}}\left({x}_0\right)=\sum_{t=1}^T \underset{q\left({x}_t \mid {x}_0\right)}{\mathbb{E}}\left\|\mu_\theta\left({x}_t, t\right)-\hat{\mu}\left({x}_t, {x}_0\right)\right\|^2,
\end{equation}
where $\hat{\mu}$ is the mean of the posterior $q({x}_{t-1}|{x}_0, {x}_t)$, and $\mu_\theta$ is the predicted mean of $p_\theta({x}_{t-1} |{x}_t)$, which is predicted by the parameterized neural models.

\section{Method: the Masked-Diffuse LM}
In this section, we describe our introduced Masked-Diffuse LM. The overall diagram is shown in Figure~\ref{fig:process}. Different from the recent diffusion models for languages, e.g., Diffusion-LM \cite{https://doi.org/10.48550/arxiv.2205.14217}, which are based on continuous diffusion models, we propose to make corruptions in both discrete and continuous space to help modeling the textual data. Specifically, we formulate a novel corruption process as an alternative to Gaussian diffusion (in Section~\ref{Sec:forward}) and we directly map continuous vectors to discrete inputs in every diffusion step with cross-entropy objectives (in Section~\ref{Sec:diffuse}). Moreover, our approach could easily integrate pre-trained language models (in Section~\ref{sec:plm}).

\subsection{Embedding}
For the input sentence $d$ with $l$ tokens $d = \hat{w}_{1:l}$, we first map the discrete tokens to the continuous space and form the initial latent variable, $X_0$, through a learnable embedding layer or an encoder $e(.)$: 
\begin{equation}
    X_0 = w_{1:l} = e(w_{1:l}).
\end{equation}
This bridges the discrete space and continuous space. We will then add designed soft-masked noise to the tokens' representations in the later diffusion models.

\subsection{Forward Process with Soft-Masking} \label{Sec:forward}
Different words in sentences play different roles. As a result, when corrupting the sentences and recovering the sentences, words with various importance should be treated differently. Thus, in this work, instead of evenly adding Gaussian noise to all the token embeddings like in Diffusion-LM \cite{https://doi.org/10.48550/arxiv.2205.14217}, we add soft-masked noise to different tokens in the input text in different stages to corrupt the text gradually with structures. Intuitively, more important words would be perturbed with soft-masks in an earlier stage so that the model could be encouraged to generate them in the later phase to follow the \textit{easy-first-generation} nature of language planning and generation.

In this work, we consider the following aspects to measure and define the importance of words in one sentence:

\paragraph{Word Relevancy} We use the tf-idf weights \cite{https://doi.org/10.5281/zenodo.4777594}, $w_{\text{tf-idf}}$, of the word as one way to measure the relevance of word $w$ in one sentence $d$:
\begin{equation}
   w_{\text{tf-idf}}(w,d) = \frac{f_{w, d}}{\sum_{w^{\prime} \in d} f_{w^{\prime}, d}} \log \frac{N}{1+|\{d \in D: w \in d\}|},
\end{equation}
where the $f_{w, d}$ is the number of times that word $w$ occurs in sentence $d$, $N$ is the number of sentences in the corpus, and $D$ is the set of sentences, and $|\{d \in D: w \in d\}|$ is number of sentences where the word $t$ appears. A higher weight for word $w$ in sentence $d$ in tf–idf means that the word might be more important in the sentence. 

\paragraph{Entropy} We also consider measuring the amount of information with entropy $H$ \cite{https://doi.org/10.48550/arxiv.1606.06996,https://doi.org/10.48550/arxiv.2211.15029} in the word $w$ to reflect the importance of that word:
\begin{equation}
    H(w) = - p\left(w\right) \log \left(p\left(w\right)\right)
\end{equation}
where $p\left(w\right)=\frac{f_w}{\sum_{j=1}^V f_j}$ represents the probability of word $w$ and $f$ is the word Reluency in the corpus. A word with lower entropy indicates that the word might contain less information and thus be less important compared to the words with higher entropy. 

In practice, we combine these two measures (with normalization) to decide the importance $I$ of the word $w$ in one sentence $d$ by:
\begin{equation}
    I(w) = \frac{x_{\text{tf-idf}}(w,d)}{\sum_{w' \in d} w_{\text{tf-idf}}(w',d)} + \frac{H(w)}{\sum_{w' \in d} H(w')}.
\end{equation}

Based on the introduced importance $I$ of the words in a sentence, we first divide these words into $m$ buckets $\{W_{1:m}\}$. The buckets with lower indices include words with higher importance. We will add soft-masked noise to words with higher importance before words with lower importance. By doing this, models could learn to generate the easier words first and then generate harder words in the reversed denoising process for better generation quality. Specifically, at every step $t$, we will add a small amount of Gaussian noise to the hidden representation of the word $w_i$ in bucket $W_{|\frac{tm}{T}|}$: 
\begin{align}
   q(w_{i,t+1}|w_{i,t}) = N(w_{i,t+1}; \sqrt{(1-\beta_t)}w_{i,t}, \beta_t I),
 \end{align} 
where $\beta_t$ is the amount of noise added at diffusion step $t$. 

We further apply a square-root noise schedule following \citet{https://doi.org/10.48550/arxiv.2205.14217} to gradually increase $\beta_t$:
\begin{equation}
    \beta_t = 1-\sqrt{t / T+s},
\end{equation}
where $s$ is a small constant that corresponds to the starting noise level. Thus, less noise would be added to harder words to stabilize the training. By performing the above noising steps, initial latent variable $X_0$ is gradually corrputed to a series of noisy latent variables $X_{1:T}$.

\subsection{Diffusion Process} \label{Sec:diffuse}
After the forward process to corrupt the input tokens in sentences $d$ into latent variables $X_{1:T}$, we then gradually denoise $X_T$ back to $X_0$ through diffusion steps, $\hat{X}_{t-1} = p(\hat{X}_{t}|\theta)$, where $\theta$ is the learned parameter to model the state transition. In practice, we model the transition with Transformers \cite{vaswani2017attention}.

After every diffusion step $t \in (0, T]$, instead of minimizing the distance between the hidden representations of $\hat{X}_{t-1}$ and $X_0$ \cite{https://doi.org/10.48550/arxiv.2205.14217}, we first directly map the continuous space to discrete space using a learnable linear layer $f(.)$ and then minimize a weighted cross entropy between the predicted sentence and (i) the original sentence $d$ and (ii) the masked sentence $\hat{d}$ at time step $t-1$:
\begin{align*}
   \mathcal{L}_{t} = \gamma_tCE(f(\hat{X}_{t-1}), d; \theta) 
   + CE(f(\hat{X}_{t-1}), \hat{d}; \theta), t \in (0, T]
\end{align*}
Here, $\gamma_t = \frac{T-t}{T}$. In other words, we put higher weights on the masked tokens that are masked in this time step during the forward process and put lower weights to the other tokens. So the models are learned to generate the corresponding masked tokens first at every time step.


\begin{table*}[t]
\centering
\caption{Main Results. The Accuracy ($\uparrow$) and the Fluency ($\downarrow$) of different methods on five controllable generation tasks including semantic content, POS, syntax tree, syntax spans and length. $\dag$ indicates our methods.} \label{Tab:main}
\vskip 0.15in
\begin{tabular}{c|cc|cc|cc|cc|cc} \toprule
{\color[HTML]{333333} }                                   & \multicolumn{2}{c}{{\color[HTML]{333333} \textbf{Semantic Content}}}                   & \multicolumn{2}{|c}{{\color[HTML]{333333} \textbf{POS}}}                                & \multicolumn{2}{|c}{{\color[HTML]{333333} \textbf{Syntax Tree}}}                        & \multicolumn{2}{|c}{{\color[HTML]{333333} \textbf{Syntax Spans}}}                       & \multicolumn{2}{|c}{{\color[HTML]{333333} \textbf{Length}}}                             \\ 
\multirow{-2}{*}{{\color[HTML]{333333} \textbf{Methods}}} & {\color[HTML]{333333} \textbf{Acc}} & {\color[HTML]{333333} \textbf{Fluency}} & {\color[HTML]{333333} \textbf{Acc}} & {\color[HTML]{333333} \textbf{Fluency}} & {\color[HTML]{333333} \textbf{Acc}} & {\color[HTML]{333333} \textbf{Fluency}} & {\color[HTML]{333333} \textbf{Acc}} & {\color[HTML]{333333} \textbf{Fluency}} & {\color[HTML]{333333} \textbf{Acc}} & {\color[HTML]{333333} \textbf{Fluency}} \\ \midrule \midrule
{\color[HTML]{333333} PPLM}                               & {\color[HTML]{333333} 9.9}                   & {\color[HTML]{333333} 5.32}             & {\color[HTML]{333333} -}                     & {\color[HTML]{333333} -}                & {\color[HTML]{333333} -}                     & {\color[HTML]{333333} -}                & {\color[HTML]{333333} -}                     & {\color[HTML]{333333} -}                & {\color[HTML]{333333} -}                     & {\color[HTML]{333333} -}                \\
{\color[HTML]{333333} FUDUGE}                             & {\color[HTML]{333333} 69.9}                  & {\color[HTML]{333333} 2.83}             & {\color[HTML]{333333} 27.0}                    & {\color[HTML]{333333} 7.96}             & {\color[HTML]{333333} 17.9}                  & {\color[HTML]{333333} 3.39}             & {\color[HTML]{333333} 54.2}                  & {\color[HTML]{333333} 4.03}             & {\color[HTML]{333333} 46.9}                  & {\color[HTML]{333333} 3.11}             \\ \midrule
{\color[HTML]{333333} Diffusion-LM}                       & {\color[HTML]{333333} 81.2}                  & {\color[HTML]{333333} 2.55}             & {\color[HTML]{333333} 90.0}                    & {\color[HTML]{333333} 5.16}             & {\color[HTML]{333333} 86.0}                    & {\color[HTML]{333333} 3.71}             & {\color[HTML]{333333} 93.8}                  & {\color[HTML]{333333} 2.53}             & {\color[HTML]{333333} 99.9}                  & {\color[HTML]{333333} 2.16}             \\
{\color[HTML]{333333}  + BERT}                & {\color[HTML]{333333} 77.4}                  & {\color[HTML]{333333} 2.68}             & {\color[HTML]{333333} 86.2}                  & {\color[HTML]{333333} 5.43}             & {\color[HTML]{333333} 82.3}                  & {\color[HTML]{333333} 3.92}             & {\color[HTML]{333333} 89.3}                  & {\color[HTML]{333333} 3.13}             & {\color[HTML]{333333} 99.9}                  & {\color[HTML]{333333} 2.68}             \\ \midrule
{\color[HTML]{333333} Masked-Diffuse LM $\dag$}                  & {\color[HTML]{333333} 81.9}                  & {\color[HTML]{333333} 2.35}             & {\color[HTML]{333333} 91.6}                    & {\color[HTML]{333333} 5.03}             & {\color[HTML]{333333} 86.6}                  & {\color[HTML]{333333} 3.66}             & {\color[HTML]{333333} 94.7}                  & {\color[HTML]{333333} 2.48}             & {\color[HTML]{333333} 99.9}                  & {\color[HTML]{333333} 2.13}             \\
{\color[HTML]{333333} + BERT $\dag$}           & {\color[HTML]{333333} \textbf{82.9}}         & {\color[HTML]{333333} \textbf{2.30}}     & {\color[HTML]{333333} \textbf{92.9}}         & {\color[HTML]{333333} \textbf{4.78}}    & {\color[HTML]{333333} \textbf{89.7}}         & {\color[HTML]{333333} \textbf{3.44}}    & {\color[HTML]{333333} \textbf{95.8}}         & {\color[HTML]{333333} \textbf{2.33}}    & {\color[HTML]{333333} \textbf{100}}          & {\color[HTML]{333333} \textbf{2.08}}    \\ \bottomrule
\end{tabular} 
\vskip -0.1in
\end{table*}

\begin{table}[t]
\small
\centering
 \caption{Training time and inference time (generating 50 samples) for different models. } \label{tab:cost}
 \vskip 0.15in
\begin{tabular}{c|c|c} \toprule
\textbf{Methods}          & \textbf{Training } (h) & \textbf{Inference} (s) \\ \midrule \midrule
Diffusion-lm      & 8.0                 & 80                 \\
+BERT             & 15.2                & 920                \\ \midrule
Masked-Diffuse LM  & 3.4                & 68                 \\
+BERT             & 4.8               & 700      \\ \bottomrule          
\end{tabular} 
\vskip -0.1in
\end{table}

\begin{table*}[t]
\centering
 \caption{The average ranking every method receives from human evaluation (lower is better). } \label{tab:human_eval}
  \vskip 0.15in
\begin{tabular}{c|c|c|c|c|c} \toprule
\textbf{Methods}           & \textbf{Semantic Content} & \textbf{POS} & \textbf{Syntax Tree} & \textbf{Syntax Spans} & \textbf{Length} \\ \midrule \midrule
Diffusion-lm      &        2.89         & 2.76    & 3.16            &   2.88           &  2.46      \\
+BERT             &        3.87          &  3.46   & 3.72            &  3.68            & 3.34        \\ \midrule
Masked-Diffuse LM &     2.56             & 2.48        &   2.88              &  2.35               &  2.18          \\
+BERT             &     \textbf{1.32}            & \textbf{1.28}    &  \textbf{1.16}           &   \textbf{1.55}           &   \textbf{1.86}    \\ \midrule
\end{tabular}
\vskip -0.1in
\end{table*}

\begin{table}[t]
\centering
\caption{Performances on Semantic Content of Masked-Diffuse LM with different types of noise applied in forward noising process. $\dag$  indicates our method.} \label{Tab:noise}
  \vskip 0.15in
\begin{tabular}{c|cc} \toprule
\multirow{2}{*}{\textbf{Noise Type}} & \multicolumn{2}{c}{\textbf{Semantic Content}} \\ 
                                     & \textbf{Acc}         & \textbf{Fluency}       \\ \midrule \midrule
Gaussian                             & 75.3                 & 3.01                   \\ \midrule
Random Mask                          & 78.8                 & 2.67                   \\
Mask w. POS                          & 80.4                 & 2.58                   \\

Mask w. Entropy &81.1 &2.44  \\

Mask w. Rel &80.8 &2.52 \\ 

Mask w. Entropy+Rel $\dag$                      & \textbf{81.6}        & \textbf{2.38}  \\ \bottomrule        
\end{tabular} 
\vskip -0.1in
\end{table}

\begin{table*}[t]
\centering
 \caption{Performances of Masked-Diffuse LM trained with different objectvies on controllable generation tasks. $\dag$  indicates our method.} \label{Tab:objective}
   \vskip 0.15in
\begin{tabular}{c|cc|cc|cc|cc|cc} \toprule
\multirow{2}{*}{\textbf{Methods}} & \multicolumn{2}{c|}{\textbf{Semantic Content}} & \multicolumn{2}{c|}{\textbf{POS}}     & \multicolumn{2}{c|}{\textbf{Syntax Tree}} & \multicolumn{2}{c|}{\textbf{Syntax Spans}} & \multicolumn{2}{c}{\textbf{Length}}    \\
                         & Acc               & fluency          & Acc           & fluency       & Acc            & fluency        & Acc             & fluency        & Acc           & fluency       \\ \midrule \midrule
L2                       & 81.1              & 2.44             & 90.6          & 5.17          & 86.2           & 3.68           & 94              & 2.51           & 99.8          & 2.14          \\
L2-BERT                  & 80.1              & 2.48             & 89.4          & 5.82          & 84.1           & 3.91           & 93.2            & 2.88           & 99.9          & 2.89          \\ \midrule
CE  $\dag$                        & {\color[HTML]{333333} 81.9}                  & {\color[HTML]{333333} 2.35}             & {\color[HTML]{333333} 91.6}                    & {\color[HTML]{333333} 5.03}             & {\color[HTML]{333333} 86.6}                  & {\color[HTML]{333333} 3.66}             & {\color[HTML]{333333} 94.7}                  & {\color[HTML]{333333} 2.48}             & {\color[HTML]{333333} 99.9}                  & {\color[HTML]{333333} 2.13}             \\
CE-BERT $\dag$                & {\color[HTML]{333333} \textbf{82.9}}         & {\color[HTML]{333333} \textbf{2.30}}     & {\color[HTML]{333333} \textbf{92.9}}         & {\color[HTML]{333333} \textbf{4.78}}    & {\color[HTML]{333333} \textbf{89.7}}         & {\color[HTML]{333333} \textbf{3.44}}    & {\color[HTML]{333333} \textbf{95.8}}         & {\color[HTML]{333333} \textbf{2.33}}    & {\color[HTML]{333333} \textbf{100}}          & {\color[HTML]{333333} \textbf{2.08}} \\ \bottomrule
\end{tabular}
\vskip -0.1in
\end{table*}

\begin{table*}[t]
\centering
\caption{Examples of the intermediate generated text of our Masked-Diffuse LM on the Length and Semantic Content tasks.} \label{Tab:case}
\vskip 0.15in
\begin{tabular}{c|l} \toprule
\textbf{Case Study}        & \multicolumn{1}{c}{\textbf{Sentences}}                                                   \\ \midrule \midrule
Input             &       \textit{7}                                                                           \\ \midrule
$t = 500$      &  [\textit{mask}] [\textit{mask}] [\textit{mask}] [\textit{mask}] [\textit{mask}] [\textit{mask}] [\textit{mask}]                                               \\
$t = 400$ &     [\textit{mask}] is an [\textit{mask}] restaurant .   \\
$t = 200$ &     The [\textit{mask}] is an Indian restaurant .  \\
$t = 0$ &     The Mill is an Indian restaurant .  \\ \midrule \midrule
Input             &   \textit{name : Travellers Rest Beefeater}                                                                              \\ \midrule
$t = 500$       &       [\textit{mask}] [\textit{mask}] [\textit{mask}] [\textit{mask}] [\textit{mask}] [\textit{mask}] [\textit{mask}] [\textit{mask}] [\textit{mask}] [\textit{mask}] [\textit{mask}] [\textit{mask}]                                                                      \\
$t = 400$ &     [\textit{mask}] Rest [\textit{mask}]  is a   [\textit{mask}] [\textit{mask}] [\textit{mask}]    that is [\textit{mask}]  .                                                              \\ 
$t = 200$ &  Travellers Rest [\textit{mask}]  is a  reasonably [\textit{mask}] restaurant   that is awesome  .  \\
$t = 0$ & Travellers Rest Beefeater is a reasonably priced restaurant that is awesome .     \\
\bottomrule                                                                           
\end{tabular}  
\vskip -0.1in
\end{table*}

\subsection{Adapting Pre-trained Language Models} \label{sec:plm}
Our introduced Masked-Diffuse LM also allows the use of large pre-trained language model \cite{devlin2018bert, liu2019roberta, yang2019xlnet, joshi2019spanbert, sun2019ernie, clark2019electra, lewis2019bart, bao2020unilmv2, he2020deberta, raffel2020exploring}. In this work, we use BERT \cite{devlin2018bert} as an example. To combine the prior knowledge in large language models, it is straightforward to directly replace the embedding layer $e(.)$ with the pre-trained model and use the pre-trained model to get the hidden representations of input tokens as the initial state in diffusion models. We use the final linear layers in pre-trained models to predict the tokens. For efficiency, in our experiments, when using pre-trained models, we freeze the parameters in them and only learn the transition model $\theta$ in our Masked-Diffuse LM.

\section{Controllable Text Generation with Masked-Diffuse LM}
In this section, we illustrate how we apply our Masked-Diffuse LM to fulfill controllable text generation. Inspired by recent plug-and-play methods \cite{yang-klein-2021-fudge,Dathathri2020Plug,krause-etal-2021-gedi-generative,liu-etal-2021-dexperts}, we conduct controls $c$ from external modules (e.g., classifiers) directly on the latent variables $X_t$ in every intermediate step $t \in [0,T]$ in our Masked-Diffuse LM:
\begin{equation}
    p\left(X_{0: T} \mid c\right)=\prod_{t=1}^T p\left(X_{t-1} \mid X_t, c\right).
\end{equation}

We follow the conditional independence assumption \cite{yang-klein-2021-fudge,Dathathri2020Plug,krause-etal-2021-gedi-generative,liu-etal-2021-dexperts} and decompose the above joint probability into a sequence of control task at every time step $t$:
\begin{equation}
\begin{aligned}
    p\left(X_{t-1} \mid X_t, c\right) &\propto p\left(X_{t-1} \mid X_t\right) \cdot p(c \mid X_{t-1}, X_t)  \\
    &= p\left(X_{t-1} \mid X_t\right) \cdot p(c \mid X_{t-1}). 
\end{aligned}
\end{equation}

As a result, for the $t$-th step, we run gradient updates on $X_{t}$ to generate $X_{t-1}$:
\begin{equation}
\begin{aligned}
    \nabla_{X_{t-1}} \log p\left(X_{t-1} \mid X_t, c\right) &= \lambda \nabla_{X_{t-1}} \log p\left(X_{t-1} \mid X_t\right) \\
&+ \nabla_{X_{t-1}} \log p\left(c \mid X_{t-1}\right),
\end{aligned}
\end{equation}
where both $\log p(X_{t-1}|X_t)$ and $\log p(c|X_{t-1})$ are differentiable: the first term is parametrized by the transition Transformers, $\theta$, in Masked-Diffuse LM, and the second term is parametrized by extra neural network classifiers. Note that the extra classifiers are trained with the diffusion latent variables as input to allow direct gradient updates on the latent space. Note that $\lambda$ is a fluency regularization hyper-parameter to balance the fluency (gradient updates from Masked-Diffuse LM) and control (gradient updates from classifiers) in order to further improve the generation quality.

For the decoding strategy, following \citet{https://doi.org/10.48550/arxiv.2205.14217}, the  Minimum Bayes Risk (MBR) decoding \cite{kumar-byrne-2004-minimum} is used to aggregate and select the sample that has the lowest expected loss under the specified loss function from the Masked-Diffuse LM.

\section{Experiments}
\subsection{Datasets}
In this work, we train our Masked-Diffuse LM on the E2E datasets \cite{novikova-etal-2017-e2e}, which consists of 50K restaurant reviews together with the labels in terms of food type, price, and customer ratings. 

Following \citet{https://doi.org/10.48550/arxiv.2205.14217}, we conduct 5 control tasks to evaluate the learned Masked-Diffuse language model: 
\begin{itemize}
    \item  \textbf{Semantic Content.} For a given field (e.g., \textit{food}) and value (e.g., \textit{Japanese}), sentences that covers field=value need to be generated. We evaluate the accuracy of the generated sentence by examine the exact match rate of ``value'' (word mention).

    \item \textbf{Parts-of-speech.} For a given sequence of parts-of-speech (POS) tags (e.g., \textit{Noun Verb Determiner Noun}), the models need to produce the sentence with the same length and follow the exact given POS tag sequence (e.g., \textit{Birds eat the warms}). We evaluate the accuracy of the generation by checking the word-level POS tag exact match (under an oracle POS tagger).
    
    \item \textbf{Syntax Tree.} For a given syntactic parse tree, the generated sentence should have the same parse tree. We evaluate the accuracy by first parsing the generated sentence with an off-the-shelf parser and report the F1 scores compared to the given parse.
    
    \item \textbf{Syntax Spans.}  For a given (span, syntactic category) pair (e.g., \textit{(2, 5, VP)}), the parse tree of the generated sentence  should match the given syntactic category over the given spans. We evaluate the accuracy of the sentence by the exact match rate of the given spans.

    \item \textbf{Length.} For a given target length (e.g., \textit{20}), the models need to generate a sentence within $\pm2$ of the given target. We evaluate the accuracy by the match rate of the sentence lengths.

\end{itemize}

For every control task, we sample 200 control targets $c$ from the validation splits, and we generate 50 samples for each control target. The first four tasks rely on a classifier to guide the diffusion, and the last one task is classifier free.  To further evaluate the fluency of the generated sentences from models, we use a teacher LM (i.e., a carefully fine-tuned GPT-2 model) and report the perplexity of generated text under the teacher LM. A lower perplexity indicates
better sample quality and fluency.

\subsection{Baselines}
We compare our Masked-Diffuse LM with the following state-of-the-art baselines on controllable generation tasks:
\begin{itemize}
    \item \textbf{PPLM} \cite{Dathathri2020Plug} runs gradient ascent on the pre-trained language models' hidden representations to increase the classifier probabilities and language model probabilities.

    \item \textbf{FUDGE} \cite{yang-klein-2021-fudge} reweights the predicted tokens from the pre-trained language models by a discriminator which takes in a prefix sequence and predicts whether the complete sequence would satisfy the constraint. 

    \item \textbf{Diffusion-LM} \cite{https://doi.org/10.48550/arxiv.2205.14217} learns an embedding to map discrete text into the continuous space
where it performs Gaussian diffusion process. Also, a rounding step is designed to map the embeddings back into discrete texts. For every control task, the Diffusion-LM infuses the controlling signals in every diffusion step.
\end{itemize}

\subsection{Experimental Setting}
We use a Transformer with 80M parameters to parameterize our Masked-Diffuse LM, with a sequence length $n = 64$, diffusion steps $T = 500$, and a square-root noise schedule. For Masked-Diffuse LM, we set the hidden dimension to $128$. We set the number of word buckets $m=3$. When combining pre-trained models, we incorporate BERT-base \cite{devlin2018bert} with about 110M parameters. We use BERT to encode the input text into vectors with dimension of $768$ and freeze the parameters in BERT. We learn Masked-Diffuse LM with the AdamW optimizer \cite{loshchilov2018decoupled} for 20,000 steps with learning rate of 3e-4, dropout probability of 0.1, and batch size of 32. We use a linear warmup schedule starting with 1,000 warmup steps. All experiments are conducted on NVIDIA A100 Tensor Core GPUs. We use 4 GPUs for training and a single GPU for sampling. 

\subsection{Results}
We show the main results on five controllable generation tasks in Table~\ref{Tab:main}. When the diffusion process is engaged, the performances on all the controlled generation tasks receives significant boosts (e.g., 81.2 of Diffusion-LM vs. 69.9 if FUDUGE on Semantic Content task), suggesting the superiority of the diffusion model on controllable generation tasks. While the previous Diffusion-LM can not be well combined with large language model like BERT (e.g., a 5\% drop on Semantic Content accuracy), largely due to the fact that their way (rounding) to bridge continuous space and discrete space suffers from significantly higher dimensions. Compared to Diffusion-LM, our proposed Masked-Diffuse LM consistently outperforms the previous models in all tasks (e.g., a 1.7\% improvement on the POS task), indicating the effectiveness of our introduced linguistic-informed noise forward process. Also, when combined with large language models like BERT, our method significantly outperforms the previous methods, demonstrating that our approach can be well aligned with pre-trained models.

\paragraph{Efficiency} We also display the training cost and inference cost in Table~\ref{tab:cost}. Compared to the previous Diffusion-LM, our method requires significantly less training time to converge and needs less inference time to generate sentences. This is because our introduced noise process is more stable and suitable for modeling languages. Besides, the objectives we introduced are more efficient than the rounding techniques in previous work.

\paragraph{Human Evaluation} We then conduct human evaluation to evaluate the generated conversations qualitatively. We ask native speakers of English from Amazon Mechanical Turk to rank the quality of 50 generated sentences (randomly sampled) from different models for every control task. Specifically, annotators need to rank different system outputs based on the (i) fluency (whether the given sentence is readable and fluent) and (ii) the controllability (whether the given sentence match the given control conditions). To increase annotation quality, we require turkers to have a 98\%
approval rate with over 10,000 approved tasks for their previous work. The pay rate was \$0.15 per hit. Every example is assessed by 3 annotators, and the rank for every sentence is aggregated by majority voting. The Intra-Class Correlation (\textit{ICC1k}) was 0.63, indicating moderate agreement \cite{koo2016guideline}. The results are shown in Table~\ref{tab:human_eval}. As it shows, our proposed Masked-Diffuse LM and its variation with BERT received the best average ranks, suggesting the effectiveness of our proposed diffusion modeling strategy for languages. 

\subsection{Ablation Studies}
We then perform ablation studies to demonstrate the effectiveness of our introduced linguistic-informed noise and the cross entropy objectives.

\paragraph{Noise Strategy} We first demonstrate the performances on Semantic Content task of Masked-Diffuse LM with different types of noise strategy in Table~\ref{Tab:noise}. \emph{Gaussian} adds Gaussian noise to all the tokens in the input sentence in the forward process following \citet{https://doi.org/10.48550/arxiv.2205.14217}. We also compare different masking noise strategies: (i) Random Mask, where the soft-mask is added to tokens in a random order. (ii) Mask with POS, where the soft-mask perturbs the tokens in an order (noun $\rightarrow$ verb $\rightarrow$ other words) based on POS tags. Our introduced noise strategy (Mask with Entropy and Reluency) shows significantly better performances on semantic content generation. This indicates that our introduced noise strategy that considers the linguistic features in sentences is providing more appropriate perturbation to the textual data for the diffusion process. 

\paragraph{Objectives} We further show the impact of different objectives in Table~\ref{Tab:objective}. We compare our used cross entropy objectives with the $L_2$ object that is used in \citet{https://doi.org/10.48550/arxiv.2205.14217} where they minimize the distance between latent intermediate variables and the initial latent variable instead of directly predicting the text. We observe that cross entropy objectives slightly perform better than $L_2$ when the pre-trained model is not used. After combining with large language models, CE-BERT significantly outperforms the $L_2$-BERT, indicating the effectiveness of our introduced objectives in terms of incorporating large language models.

\subsection{Case Studies}
We also include some examples of intermediate steps of Masked-Diffuse LM in Table~\ref{Tab:case}. In the denoising diffusion process, easy words are generated first. For example, ``\textit{is}'', ``\textit{an}'', and ``\textit{restaurant}''. With more diffusion steps, sentences are enriched with more informative words such as ``\textit{Mill}'' and ``\textit{Indian}''. It shows that our Masked-Diffuse LM encourages the generation to follow an easy-first order for stable and better generation quality.

\section{Conclusion}
In this work, we present a novel diffusion model for language, Masked-Diffuse LM, which corrupts the discrete text with a linguistic-informed soft-masking strategy and then iteratively denoises them back by directly predicting the text. Specifically, we gradually soft-mask the tokens in the sentence following an order from more informative words to less informative words in the forward process. This satisfies the flexibility for diffusion models, as well as encourages the easy-first-generation nature in the denoising process for better generation quality. Also, we directly predict the discrete token during the diffusion process with the cross-entropy loss to stabilize the intermediate diffusion steps and make our approach orthogonal to large pre-trained language models. Experiments on E2E dataset and five controllable generation tasks including Semantic Content, Parts-of-speech, Syntax Tree, Syntax Spans, and Length show that our Masked-Diffuse LM can (i) achieve the state-of-the-art performances compared to recent baseline models and (ii) allow more efficient training and inference compared to the previous Diffusion-LM.

\bibliography{draft}
\bibliographystyle{unsrtnat}

\end{document}